\documentclass[runningheads]{llncs}
\usepackage[T1]{fontenc}
\usepackage{graphicx}
\usepackage{newtxmath}

\usepackage{amsmath}
\usepackage{booktabs}
\usepackage{tikz}
\usetikzlibrary{shapes.geometric, arrows}
\usetikzlibrary{positioning,arrows.meta,calc} 
\begin{document}
\title{PDSE: A Multiple Lesion Detector for CT Images using PANet and Deformable Squeeze-and-Excitation Block}
%
%

%
%

\author{Di Fan\inst{1}\orcidID{0009-0005-8044-9063} \and
Heng Yu\inst{2}\orcidID{0000-0001-9205-2358} \and
Zhiyuan Xu\inst{1}\orcidID{0009-0005-2426-0813}}
\authorrunning{Di et al.}
%
\institute{
University of Southern California, Los Angeles, CA 90089, USA \and
Carnegie Mellon University, Pittsburgh, PA 15213, USA \
}
\maketitle              
\begin{abstract}
Detecting lesions in Computed Tomography (CT) scans is a challenging task in medical image processing due to the diverse types, sizes, and locations of lesions. Recently, various one-stage and two-stage framework networks have been developed to focus on lesion localization. We introduce a one-stage lesion detection framework, PDSE, by redesigning Retinanet to achieve higher accuracy and efficiency for detecting lesions in multimodal CT images. Specifically, we enhance the path aggregation flow by incorporating a low-level feature map. Additionally, to improve model representation, we utilize the adaptive Squeeze-and-Excitation (SE) block and integrate channel feature map attention. This approach has resulted in achieving new state-of-the-art performance. Our method significantly improves the detection of small and multiscaled objects. When evaluated against other advanced algorithms on the public DeepLesion benchmark, our algorithm achieved an mAP of over 0.20.
\keywords{Lesion Detection, RestinaNet, Deformable, Squeeze-and-Excitation.}
\end{abstract}
\section{Introduction}
Accurate lesion detection in Computed Tomography (CT) images is crucial for clinical applications. Precise initial location of the lesion can greatly facilitate subsequent processing procedures for radiologists \cite{jin2021predicting,waite2019analysis,jiang2021development}. Unlike organ localization, lesion detection is more challenging due to the diverse shapes and sizes of lesions. Different patients may present with various lesion appearances within the same organ, and different organs can show a range of symptoms for different diseases. The small size of some lesions further complicates their detection. Consequently, there is a significant demand for computer-aided methods to assist radiologists and physicians in identifying lesion localization in CT images.

With the rapid advancement of deep learning, researchers have begun to adopt this innovative approach to the detection of lesions \cite{jiang2023review,henry2020mixmodule,yu2022gpu}. However, most efforts have been concentrated on detecting specific types of lesions in CT images, such as tuberculosis lesion in lung \cite{kalinovsky2017lesion,ting2019development,chang2024synchronization}, lung nodule \cite{huang2017lung,van2024comparison}, liver lesion \cite{huang2017lung,ben2016fully,ying2024multicenter}, brain lesion \cite{pawlowski2018unsupervised,iwamura2024thin} etc. However, lesions commonly occur in different organs, making it essential to develop a model that can mimic the ability of radiologists to recognize various lesions in multiple organs. Our goal is to create a model capable of analyzing lesions in multiple organs to provide an initial estimation of lesion localization, thus enhancing the overall diagnostic process. \par

The DeepLesion dataset \cite{yan2018deeplesion} is currently the largest repository, containing over 32K CT images with body part lesions annotated with measurements and 2D bounding boxes. Several studies have utilized this public dataset. In the realm of two-stage detection methods, 3DCE \cite{yan20183d} incorporates 3D context into 2D region-based CNNs, outperforming Faster R-CNN \cite{ren2015faster}. However, its reliance on 3D context information only from the last convolution layer reduces its sensitivity to small lesions, thereby decreasing the accuracy in detecting early-stage diseases. ULDOR \cite{tang2019uldor} employs 2D Mask R-CNN \cite{he2017mask} to construct pseudo masks for each lesion region and introduces a hard negative example mining strategy to minimize false positives. Nonetheless, their assumption of elliptical lesion geometry can produce inaccurate pseudo masks, affecting dense supervision efficacy. MULAN \cite{yan2019mulan}, based on an enhanced Mask R-CNN framework with three head branches and a 3D feature fusion strategy, also has limitations due to its training dataset comprising only 22K images, leading to potential false positives in lesion detection. For one-stage methods, an improved RetinaNet \cite{zlocha2019improving} with optimized anchors and weak dense masks from weak RECIST labels has been proposed. This method, however, uses fake labels for semi-supervised learning, which can introduce noise into the model.\par
To achieve a higher detection rate, we propose a one-stage detector that does not require a region proposal network for lesion evaluation. To enhance small lesion detection, we adopt RetinaNet \cite{lin2017focal} for its high-speed, high-accuracy detection capabilities. We enhance the path aggregation flow with low-level feature maps to densify low-level features. To further improve model capabilities, we integrate adaptive deformable squeeze-and-excitation (SE) blocks and channel feature map attention to enrich feature layer expressiveness \cite{hu2018squeeze,song2024plu}. The deformable SE blocks adapt to changes in scale and fuse local information, while simultaneously modeling global information to integrate local details into the global context. This approach has led to new state-of-the-art performance. Compared to other existing methods, our approach significantly improves the detection of small and multiscaled objects, achieving a 0.20 mAP for detecting all nine types of lesions, demonstrating its superiority over other algorithms.\par
\section{Methodology }

\begin{figure}
\centering
 {\includegraphics[width=0.9\textwidth]{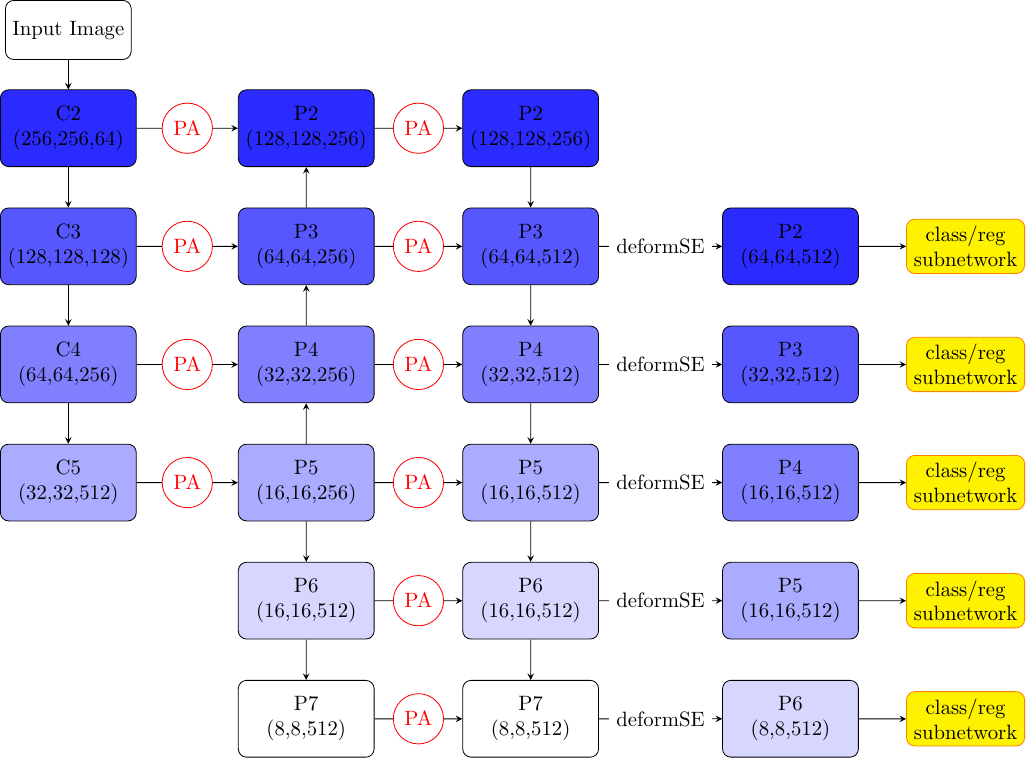}}
\caption{Framework structure for PDSE, where PA is the Path Aggragation Net (PANet) and deformSE is Deformable Squeeze-and-Excitation (DSE) block.}
\label{fig:1}
\end{figure}

Our model (Figure \ref{fig:1}) is based on  RetinaNet, Path Aggragation Net (PANet) and deformable SE block three parts. The first column is the ResNet, the second column, the third column and fourth column are feature pramid Nets (FPN), the fifth column is class and box subnets for classify and regression. We connect the first column and second column networks with PANet, the second column and third column networks with PANet, the third column network and the fourth column networks with deformable SE blocks. We first describe how path aggregation improves RetinaNet. We then show the  mechanism for further improving detection performance with Deformable Squeeze-and-Excitation (DSE) block.
\subsection{Improved RetinaNet with path aggregation}

Our model backbone is RetinaNet, a powerful one-stage detection framework that utilizes focal loss to address classification imbalance and improve accuracy in one-stage detectors. RetinaNet's architecture combines ResNet \cite{he2016deep} and Feature Pyramid Network (FPN) \cite{lin2017feature}, employing class and box subnetworks at each feature map of the FPN for classification and regression. This design allows for the detection of objects of various sizes. In our implementation, we utilize ResNet-50 as the backbone for the network structure.

Path Aggregation Network (PANet) \cite{liu2018path} aims to enhance the propagation of information flow within the framework of instance segmentation based on candidate regions. The top-down path in PANet improves the positioning information flow in the lower layers and shortens the information path between lower-level and higher-level features, thus enhancing the entire feature hierarchy. Additionally, PANet introduces adaptive feature pooling, which connects the feature grid and all feature levels, allowing the useful information from each feature level to be directly propagated to the subsequent candidate regional subnetworks. In our design, we integrate PANet between ResNet and FPN, as well as between FPN layers, to improve the accuracy of the information flow.

\subsection{Deformable Squeeze-and-Excitation (DSE) block}
\begin{figure}
\centering
 {\includegraphics[width=1\textwidth]{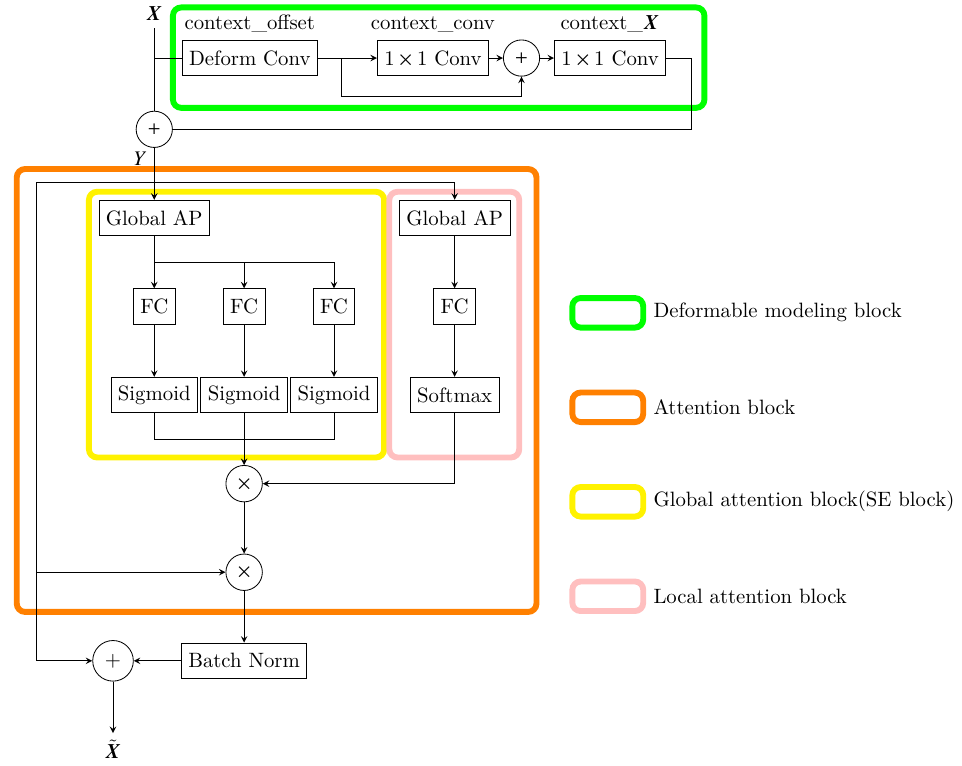}}
\caption{Framework for PDSE.}
\label{fig:2}
\end{figure}

In Figure \ref{fig:2}, we illustrate the structure of the Deformable Squeeze-and-Excitation (DSE) block. The DSE block primarily consists of two components: the Deformable Modeling block and the Attention block.\par
The Deformable Modeling block includes the Deformable Convolutional Neural Network (Deform CNN) and two additional $1 \times 1$ bottleneck CNNs. Traditional CNNs are limited in their geometric transformation capabilities due to their fixed geometry. To better approximate the shape and size of actual lesions, deformable convolution \cite{dai2017deformable} is employed. This approach introduces a 2D offset to the regular sampling grid positions of standard convolution, allowing the sampling grid to be freely deformed and better adapt to the geometry of the lesions.\par
The Attention block comprises two main components: the Global Attention block and the Local Attention block. The Global Attention block is based on the Squeeze-and-Excitation (SE) model. The core idea of the SE model is to learn feature weights according to the loss through the network, assigning larger weights to effective feature maps and smaller weights to less effective or irrelevant feature maps, thereby improving model performance. The SE block consists of three steps: Squeeze, Excitation, and Reweight.
\begin{itemize}
    \item Squeeze: This operation compresses the features along the spatial dimensions, transforming each two-dimensional feature channel into a single real number. This real number captures the global distribution of responses on a feature channel, providing a global receptive field and matching the number of input feature channels.
    \item Excitation: Similar to a gate in a recurrent neural network, this operation generates weights for each feature channel using a learned parameter \textit{w}. The parameter 
\textit{w} explicitly models the correlation between feature channels, enabling the network to focus on the most informative features.
\item Reweight: The generated weights are used to reweight the feature maps, emphasizing the most significant features while suppressing less important ones, thereby enhancing the overall feature representation and improving the model's performance.
\end{itemize}
The Reweight operation uses the weights obtained from the Excitation step to evaluate the importance of each feature channel. These weights are then applied to the original features through channel-wise multiplication, effectively recalibrating the original features along the channel dimension. The Global Attention block and the Local Attention block work together to fuse local information into the global context, enriching the entire feature layer and enhancing the model's ability to recognize the target.
\section{Experiment}
\subsection{Datasets, preprocessing and augmentation}
The NIH DeepLesion benchmark is a public dataset containing 32,120 axial CT slices (mostly 512x512) from 10,594 studies of 4,427 unique patients. Each slice contains 1 to 3 lesions, totaling 32,735 lesions. Unlike existing datasets that focus on specific types of lesions, DeepLesion classifies lesions into eight categories: lung, abdomen, mediastinum, liver, pelvis, soft tissue, kidney, and bone. The lesions are annotated with 2D bounding boxes and RECIST diameters on the key slices. \par
For pre-processing, we subtract 32,768 from each pixel intensity to obtain the Hounsfield Units (HU) values. We then clip these values into the range of -1024 to 3071 HU, covering the intensity ranges for lung, soft tissue, and bone. The public DeepLesion dataset is split into 70\% for training, 15\% for validation, and 15\% for testing.\par

For data augmentation, we implement a strategy where we divide and shrink five areas of the image to better characterize local area information. By selecting only the samples that contain the target object, we can balance the positive and negative samples more effectively.
\subsection{Quantitative Results}
\begin{table}[ht]
\centering
\caption{mAP(mean Average Precision) for different algorithms comparison}
\label{tab:1}
\begin{tabular}{@{}lcccccc@{}}
\toprule
Lesion type     & RetinaNet & RetinaNet-PA & Ours & Faster R-CNN & CenterNet & FSAF \\ \midrule
Bone            & 0.1279    & 0.1398       & \textbf{0.1476 }          & 0.0542      & 0.0549    & 0.0422 \\
Abdomen         & \textbf{0.0974}    & 0.0952       & 0.0963           & 0.0453      & 0.0432    & 0.0727 \\
Mediastinum     & \textbf{0.1928}    & 0.1803       & 0.1859           & 0.0780      & 0.0763    & 0.0681 \\
Liver           & 0.1560    & 0.1707       & \textbf{0.1771}           & 0.0709      & 0.0694    & 0.0800 \\
Lung            & 0.2365    & \textbf{0.2466}       & 0.2456          & 0.1030      & 0.0730    & 0.0849 \\
Kidney          & 0.2219    & 0.2573       & \textbf{0.2746}           & 0.1150      & 0.0821    & 0.0766 \\
Tissue          & 0.1444    & 0.2075       & \textbf{0.2118}           & 0.0869      & 0.0670    & 0.0880 \\
Pelvis          & 0.1121    & \textbf{0.1162}       & 0.1066           & 0.0410      & 0.0565    & 0.0708 \\
Other           & 0.3630    & 0.3692       & \textbf{0.3831}           & 0.1443      & 0.1251    & 0.1390 \\ \bottomrule
\end{tabular}
\end{table}

To demonstrate the benefits of our model, we compare it with the following methods: (1) Faster R-CNN; (2) ULDOR; (3) RetinaNet; (4) RetinaNet+PANet; (5) RetinaNet+PANet+DSE blocks; (6) CenterNet \cite{duan2019centernet}; and (7) FSAF \cite{zhu2019feature}. Faster R-CNN is a widely used traditional detection method, while ULDOR is based on Mask R-CNN. The quantitative results are displayed in Table \ref{tab:1}. As shown, RetinaNet achieves a higher mAP than Faster R-CNN and ULDOR, indicating the superiority of this one-stage detector over the other two-stage detectors. The focal loss in RetinaNet significantly contributes to the high detection rate by addressing class imbalance. For the nine classifications in the dataset, we present the mAP results for eight specific types of lesions. Among all methods, the highest detection mAP is achieved for kidney lesions, followed by lung lesions. These two types of lesions are easier to detect due to their typically more obvious features. Conversely, the detection results for abdominal lesions are relatively low, as abdominal lesions in this dataset are more complex compared to others.
Additionally, we compare two anchor-free networks, CenterNet and FSAF, with other anchor-based networks. The results indicate that on this dataset, the anchor-free CNN models do not show an advantage over anchor-based CNN models. Although FSAF achieves relatively higher results due to its more detailed structure, its performance is still suboptimal. CenterNet, in particular, highlights the issue with anchor-free models, achieving a low mAP of 0.0718. The diverse and especially smaller-sized lesion types in this dataset reveal the limitations of the anchor-free mechanism, showing no significant benefit in lesion detection.
\subsection{Qualitative Results}
\begin{figure}
  \includegraphics[width=1.1\linewidth]{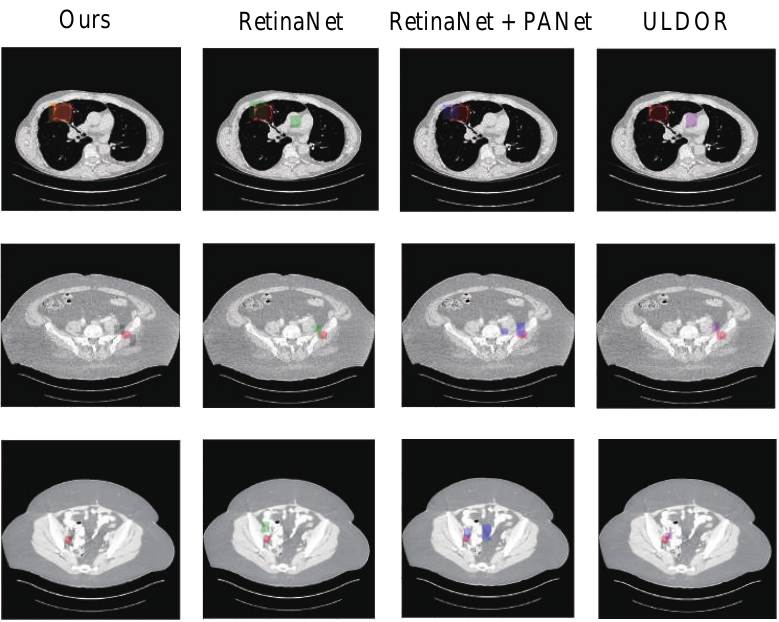}
  \caption{Detection results}
  \label{fig:3}
 \end{figure}

Figure \ref{fig:3} illustrates the nine types of lesions and the qualitative lesion detection results produced by the models trained with the configurations mentioned above. In the figure, the red box denotes the ground truth, the green box represents RetinaNet, the blue box represents RetinaNet + PANet, the black box represents RetinaNet + PANet + DSE, and the purple box represents ULDOR. From the figure, it is evident that our method (represented by the black box) provides much more accurate bounding boxes for detecting lesions, closely matching the ground truth.

 \subsection{Ablation Studies}

 We further investigate the results of our baseline RetinaNet with its default configuration, with PANet, and with both PANet and DSE. RetinaNet + PANet achieves a higher mAP than RetinaNet alone, indicating that PANet helps RetinaNet learn more discriminative features, thereby improving detection performance. The highest mAP of 0.2053 is achieved by RetinaNet + PANet + DSE, demonstrating that this combination is particularly powerful for lesion detection.

This improvement is attributed to several factors. PANet enhances the density of shallow features, making them more suitable for detecting small targets. We also integrated PANet between each layer in ResNet and FPN, and between FPN layers, which significantly improves detection accuracy. The DSE block further refines the information by ensuring that the features learned are closer to the actual lesion shapes. The deformable convolution adapts the shape to better match the lesions, while the Squeeze-and-Excitation mechanism enriches the feature map, enhancing its expressive capability.
\section{Conclusions and Future Work}
In this paper, we introduce a model that integrates PANet and the deformable SE block with the RetinaNet backbone for CT image detection. Our approach demonstrates promising results in the specific task of lesion detection within this dataset. This end-to-end, easy-to-implement structure significantly improves detection accuracy, particularly for smaller lesions. With a mAP of 0.2053, our model outperforms other detectors for 2D image inputs. In the future, we aim to adapt this model to a wider range of datasets and enhance its domain adaptation capabilities.

%
%
\bibliographystyle{splncs04}
%
\bibliography{bibliography}
\end{document}